# AI advice suppresses people's willingness to say "I don't know", even when the advice is wrong and accuracy is incentivized

Chiara Marcoccia[1], Walter Quattrociocchi[2], Valerio Capraro[3]

[1] Ecole Normale Supérieure, Paris, France

[2] University of Rome "La Sapienza", Rome, Italy

[3] University of Milan-Bicocca, Milan, Italy

Contact author: valerio.capraro@unimib.it

**Abstract**

Knowing when to say "I don't know" is fundamental to human judgment, yet AI assistants offer a fluent answer to almost any question. In five experiments (N = 3,132; four preregistered, one direct replication), participants answered difficult questions and could always decline to respond. We engineered the questions so that AI advice was wrong, separating AI use from its accuracy. Merely having access to AI nearly eliminated participants' willingness to suspend judgment, and this held whether the advice was actively requested or simply displayed. Consequently, participants answered more questions but were correct about a third as often as when AI was unavailable—yet their confidence nearly doubled. Incentivizing accuracy and penalizing inaccuracy led participants to seek and follow AI advice less, answer more accurately, and suspend judgment more often, though still far less than when AI was unavailable. As AI suggestions grow ubiquitous and unsolicited, they may not simply affect answer accuracy; they may even alter the metacognitive threshold at which people decide whether they know enough to answer.

## Introduction

Human judgment involves not only selecting an answer, but also deciding whether one possesses sufficient knowledge to answer at all. This metacognitive decision is fundamental whenever uncertainty cannot be resolved with confidence (Ackerman & Thompson, 2017). The capacity to suspend judgment under uncertainty, to say “I don’t know” rather than commit to a guess, protects people from acting on false beliefs and is a precondition for calibrated reasoning (Friedman, 2013; Lichtenstein & Fischhoff, 1977). Large language models (LLMs) can now answer almost any question instantly and fluently, and they are being woven into many tasks where this capacity matters, including searching for information, forming hypotheses, and evaluating claims (Capraro et al., 2024; Glickman & Sharot, 2024; Messeri & Crockett, 2024). Yet, these models are notoriously reluctant to admit uncertainty, even when facing a knowledge gap (Jones 2024; Zhao 2025). This raises an unresolved empirical question: when humans can seek advice from AI, does the mere availability of an answer change the decision of whether to answer at all under uncertainty?

Concern that people over-rely on AI is by now widespread, and research has examined many of its facets, from cognitive offloading (Risko & Gilbert 2016; Dimant, 2026; Poquet et al. 2026; Shaw & Nave, 2026), deskilling (Stadler et al. 2024; Lee et al. 2025; Gerlich 2025; Rossi et al. 2026), and the erosion of human agency (Kay et al. 2024; Barry & Stephenson 2025; Di Plinio 2025; Capraro, 2026). The worry is sharpened by evidence that LLMs are exceptionally persuasive (Costello et al., 2024; Matz et al. 2024; Salvi et al. 2025; Hackenburg et al. 2026), even when conveying misinformation (Spitale et al. 2023; Schoenegger et al. 2025). Neural evidence also suggests that LLMs assistance alters the cognitive activity people bring to a task (Kosmyna et al. 2025). We focus on a different facet: the willingness to suspend judgment, to withhold an answer under uncertainty. Rather than asking whether AI changes what people answer, we ask whether it changes the criterion people use to decide whether they should answer in the first place.

Previous literature already suggests that engaging with an LLM may lead people to override judgment. Shaw and Nave’s (2026) notion of “cognitive surrender” comes closest to our concern. The authors conducted a series of experiments that covertly manipulated AI accuracy. Participants who consulted an erring AI became less accurate yet more confident, even after errors, and neither incentives nor time pressure eliminated the pattern. Crucially, however, these studies measure whether people end up correct or incorrect; they do not measure whether people withhold judgment at all. Declining to commit and saying “I don’t know” was not a measured outcome. Whether the availability of AI changes people’s willingness to suspend judgment under uncertainty therefore remains untested. We study precisely this previously unexamined notion.

A central obstacle to studying AI reliance is that following AI advice is usually *rational*. When the advice is good, deferring to it is sensible, and any decline in independent judgment is difficult to interpret as a cognitive cost rather than an efficient division of labor. We remove this confound by design. Our questions ask about fine visual details in films. For example, the color of the team's uniform in *Bend It Like Beckham*, Agatha's signature hairstyle in *The Grand Budapest Hotel*, or the vehicle Monica drives in *Like a Cat on a Highway*. We chose details minor enough to be largely absent from the text available online that are therefore likely to induce hallucinations in language models largely trained on online text. The questions span a deliberate range, from superficial details in widely seen films to lesser-known titles (for instance, the color of the turtle in Enzo d'Alò's animated film *Momo*). The LLM used in our experiments (Step 3.5 Flash) answered such questions incorrectly almost without exception. We also checked some state-of-the-art LLMs (GPT-5.5, Claude 4.6 Sonnet, Gemini 3.5 Flash); they all failed on the hardest question (Monica's vehicle), while being frequently correct on the other questions (see Supplementary Information, Section S1.1). Importantly, our goal is not to estimate the overall accuracy of AI assistants, but to isolate the behavioral effect of making an answer available when that answer is systematically unreliable. Therefore, we use Step 3.5 Flash. Because the advice participants could obtain was usually wrong, any reduction in judgment suspension cannot be explained as sensible delegation to a reliable tool. It instead reveals a tendency to defer to AI output. And potentially, it would indicate that the mere availability of a plausible answer changes how people regulate uncertainty. This allows us to disentangle, within a single design, how AI access impacts both accuracy and judgment suspension.

Across five experiments (total $N = 3{,}132$; four pre-registered, one direct replication), participants answered six difficult movie questions and could always decline to answer. We first compared judgment suspension between participants who could consult an LLM and those who could not (Studies 1a–1b). Because withholding an answer should matter most when errors carry a cost, Studies 2–4 then crossed AI access with monetary stakes for accuracy, testing whether monetary consequences could restore suspension, and whether AI availability would blunt their effect; Study 2 additionally measured participants' confidence in their answers. We reasoned that monetary stakes should raise the value of admitting ignorance and therefore increase judgment suspension, but that this corrective effect might be blunted when AI advice was available, because the pull to defer to a fluent answer might override the incentive to withhold. We pre-registered this expectation as a negative interaction between AI availability and stakes on suspension. As we report below, the data did not bear this out: AI access and stakes instead operated largely independently, and the one place they interacted was accuracy rather than suspension. Finally, because AI advice increasingly reaches people unsolicited (through search summaries, writing assistants, and autocomplete), Study 4 removed participants' control over whether advice appeared, presenting it automatically to test whether the effect survives when the AI is never actively consulted.

The results are consistent and stark. Merely having access to AI advice nearly eliminated participants' willingness to suspend judgment, even though the advice was usually wrong. As a result, participants answered more questions but were correct less often: AI access made people more assured and less accurate. Monetary incentives for accuracy reduced how often participants sought and followed AI advice and improved their accuracy, and modestly increased suspension, though it remained far below the level seen when AI was unavailable. Contrary to our pre-registered prediction, incentives did not do this by attenuating the effect of AI on suspension: AI availability and stakes acted largely independently, and the only place they interacted was accuracy, where incentives helped specifically when AI was present. Together, these findings suggest that AI assistance influences not only the answers people give, but also the metacognitive decision of whether to answer under uncertainty.

**Results**

In Study 1a, participants (N = 314) answered six difficult questions about movies and were randomly assigned either to a baseline condition or to a condition in which, for each question, they could decide whether to seek advice from an AI tool. See *Methods* for details.

Participants in the baseline condition were substantially more likely to suspend judgment than participants who could seek AI advice (0.36 vs. 0.06; one-tailed t-test: $t = 8.295$, $p < .001$; see *Figure 1a*). Thus, merely making AI advice available dramatically reduced participants' tendency to withhold an answer under uncertainty, despite the fact that they remained free to decline answering.

One potential limitation of Study 1a is that the AI tool failed to respond in approximately 10% of requests, possibly due to overload or malfunctioning. Although this frequency was too small to plausibly account for the large effect, we conducted a replication designed to eliminate this concern.

In Study 1b (N = 310), we pre-generated three incorrect AI responses for each question. Participants in the AI condition who chose to seek advice were randomly shown one of these responses. See *Methods* for details.

The results replicated those of Study 1a. Participants in the baseline condition were again substantially more likely to suspend judgment than participants who could seek AI advice (0.44 vs. 0.03; $t = 11.772$, $p < .001$; see *Figure 1b*). See *Table S1* for statistical details. The replication therefore rules out transient failures of the AI system as a potential explanation for the observed effect.

Taken together, Studies 1a–1b provide initial evidence that the mere possibility of consulting AI advice nearly eliminates people's willingness to suspend judgment. Importantly, this effect emerges before introducing monetary incentives or unsolicited AI advice, suggesting that the availability of an AI-generated answer alone is sufficient to alter participants' decision of whether to answer under uncertainty.

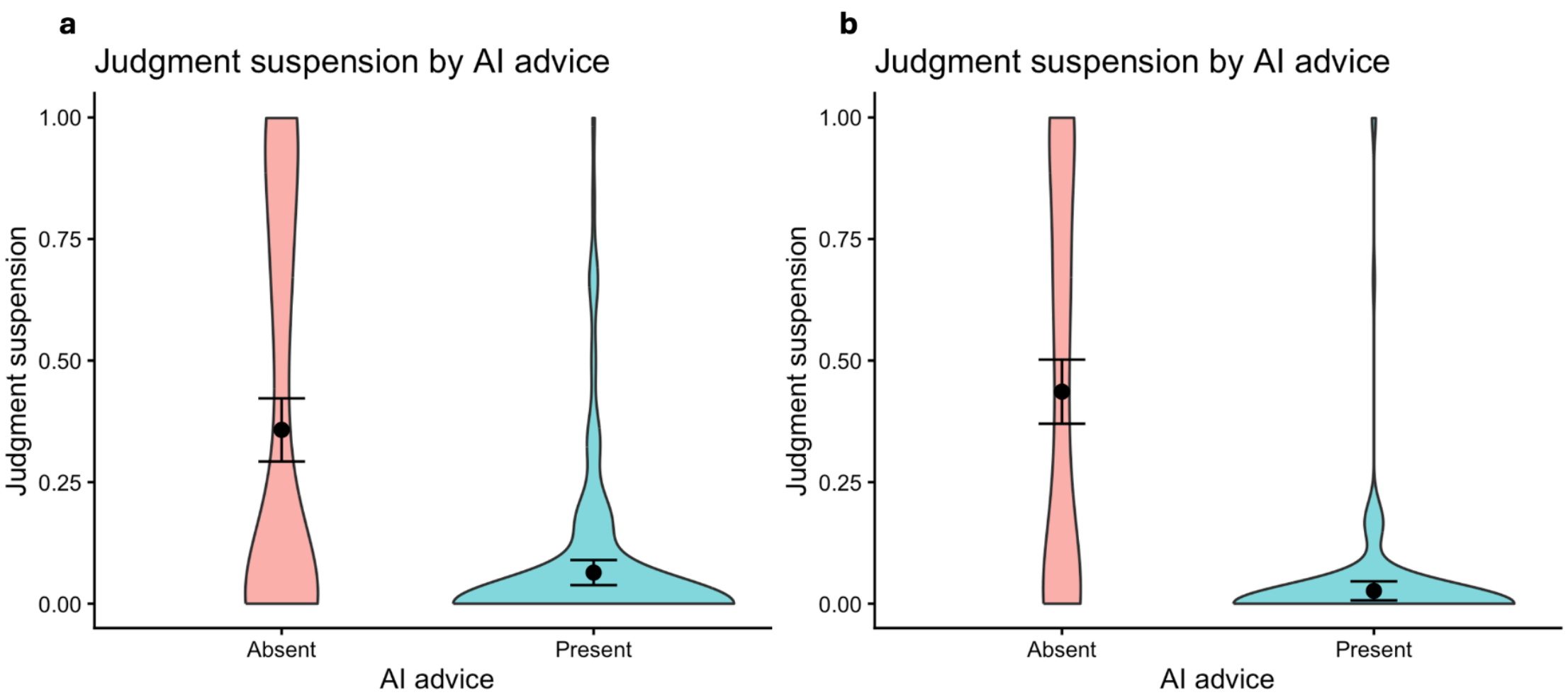

**Fig. 1 | Judgment suspension across conditions in Studies 1a and 1b.** The possibility to seek advice from an AI tool nearly eliminated people's willingness to suspend judgment. This effect emerged both in Study 1a, in which participants could query a live AI system (**a**), and in Study 1b, in which AI responses were pre-generated to eliminate possible tool malfunctions (**b**). Points represent condition means and error bars indicate 95% confidence intervals.

In these first studies, however, little was at stake: venturing a wrong answer cost no more than withholding one. Yet when errors are costly the willingness to suspend judgment should matter more. People are known to regulate whether to volunteer or withhold an answer by adjusting a confidence criterion in line with the payoffs for accuracy, withholding more as the cost of error rises (Koriat & Goldsmith, 1996), and incentives more broadly raise the effort and care people devote to a task (Camerer & Hogarth, 1999). This raises the question of whether the near-elimination of judgment suspension under AI access is a shallow default that real consequences would correct, or a more robust deferral to the AI that persists even when accuracy is rewarded.

In Study 2, participants (N = 812) answered six difficult questions about movies and were randomly assigned to one of four conditions in a 2 (AI advice: absent vs. present) × 2 (stakes: absent vs. present) design. In the stake conditions, participants earned $0.10 for each correct answer, lost $0.10 for each incorrect answer, and received $0 for suspending judgment. After each response, participants also reported their confidence level. See *Methods* for details.

Consistent with Studies 1a–1b, access to AI advice nearly eliminated judgment suspension in the absence of monetary stakes (0.17 vs. 0.01; linear regression: b = -0.159, SE = 0.023, t = -6.810, p $< .001$). The same effect was observed when monetary stakes were present (0.21 vs. 0.02; $b = -0.193$, $SE = 0.024$, $t = -7.958$, $p < .001$). Monetary stakes did not significantly increase judgment suspension either in the absence of AI advice (0.17 vs. 0.21; $b = 0.038$, $SE = 0.033$, $t = 1.145$, $p = .253$) or when AI advice was available (0.01 vs. 0.02; $b = 0.004$, $SE = 0.010$, $p = .700$). Our pre-registered primary hypothesis was that stakes would increase suspension but that AI availability would blunt this effect, yielding a negative AI × stakes interaction. This interaction was not significant ($b = -0.034$, $SE = 0.034$, $t = -1.011$, $p = .312$; see *Figure 2a*). See *Tables S2-S3* for statistical details. Thus, monetary consequences did not restore suspension, and did not moderate the effect of AI availability on it.

Confidence judgments revealed a large main effect of AI advice, which nearly doubled reported confidence (29.6 vs. 75.9 without stakes; 37.7 vs. 73.5 with stakes), alongside a significant interaction between stakes and AI advice ($b = -10.509$, $SE = 3.814$, $t = -2.756$, $p = .006$). Monetary stakes increased confidence when participants could not seek AI advice (29.6 vs. 37.7; $b = 8.083$, $SE = 3.163$, $t = 2.556$, $p = .011$), but this effect disappeared when AI advice was available (75.9 vs. 73.5; $b = -2.426$, $SE = 2.216$, $t = -1.095$, $p = .274$; see *Figure 2b*). See *Tables S4-S5* for statistical details. Thus, AI advice strongly increased confidence both with and without stakes, with a larger increase in the absence of stakes.

To better understand how incentives influenced behavior, we conducted exploratory analyses of answer correctness. Monetary stakes increased correctness both when AI advice was unavailable (0.28 vs. 0.33; $b = 0.058$, $SE = 0.021$, $t = 2.836$, $p = .005$) and when AI advice was available (0.10 vs. 0.17; $b = 0.073$, $SE = 0.017$, $t = 4.166$, $p < .001$). The interaction between stakes and AI advice was not significant ($b = 0.015$, $SE = 0.027$, $t = 0.560$, $p = .576$; see *Figure 2c*). See *Tables S6-S7* for statistical details.

Because the AI advice was identical across stake conditions, the gain in correctness in the stake condition cannot be attributed to differences in the quality of the advice itself. Instead, one possibility is that monetary incentives changed participants' willingness to rely on AI advice. Consistent with this interpretation, participants sought AI advice less frequently when monetary stakes were present than when they were absent (4.93 vs. 5.44 times out of six questions; $b = -0.503$, $SE = 0.160$, $t = -3.147$, $p = .002$; see *Figure 2d*). See *Table S8* for regression details. This behavioral change provides a plausible explanation for why incentives improved accuracy without restoring judgment suspension.

These findings provide preliminary evidence that, although incentives do not restore judgment suspension to the level observed when AI is unavailable, they nonetheless reduce reliance on AI advice and improve the accuracy of participants' answers.

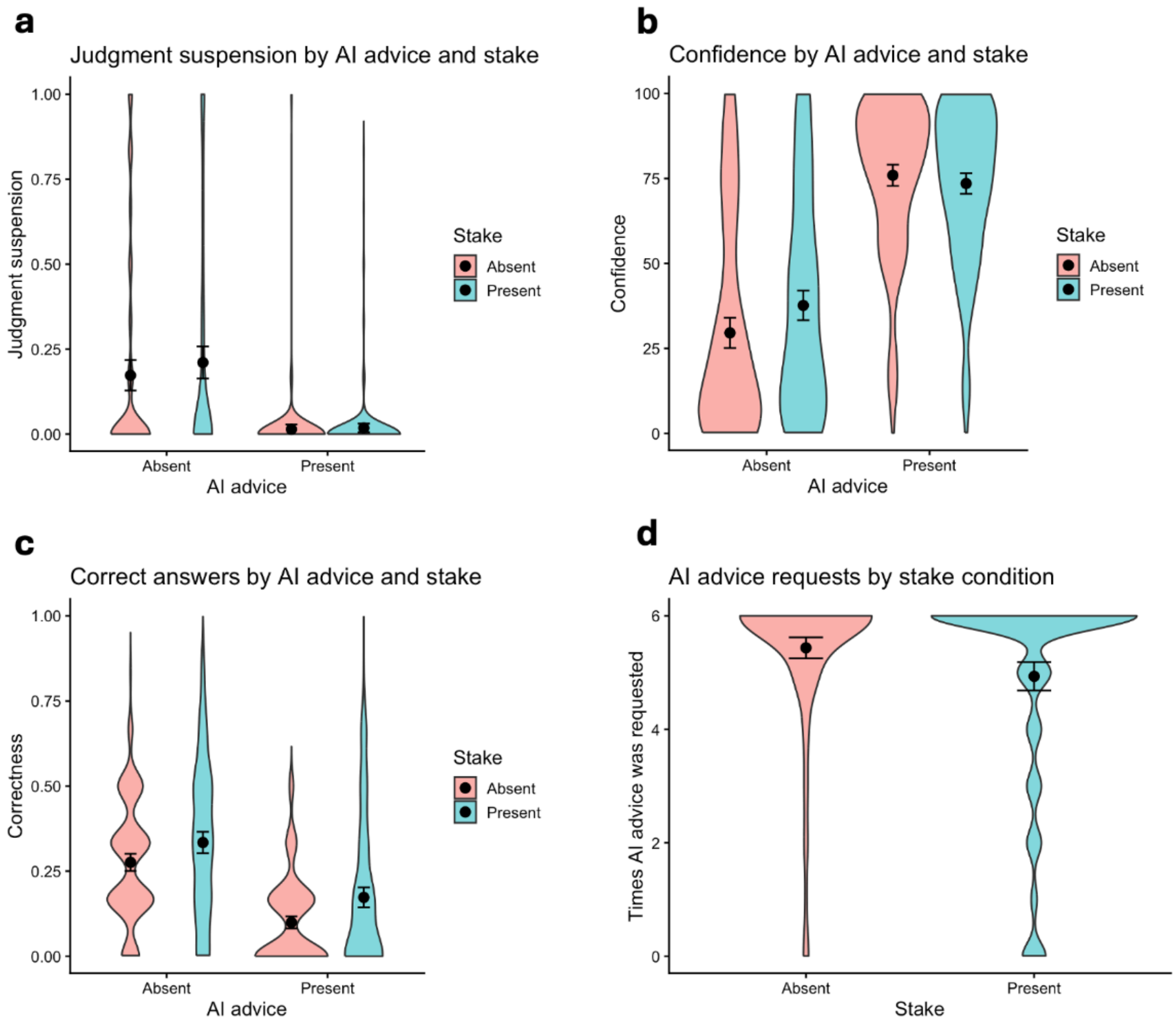

**Fig. 2 | Effects of AI advice and monetary stakes on judgment suspension, confidence, correctness, and AI-advice requests in Study 2.** Access to AI advice nearly eliminated judgment suspension both in the absence and presence of monetary stakes, with no significant interaction between the two factors (**a**). Monetary stakes increased confidence when AI advice was absent, but this effect disappeared when AI advice was available, yielding a significant interaction between stakes and AI advice (**b**). Access to AI advice reduced answer correctness both with and without monetary stakes, with no significant interaction between the two factors (**c**). Participants requested AI advice less frequently when monetary stakes were present (**d**). Points represent condition means and error bars indicate 95% confidence intervals.

Study 2 has two potential limitations. First, judgment suspension in the no-AI condition was substantially lower than in Study 1, raising the possibility that eliciting confidence judgments encouraged participants to guess rather than suspend judgment. Second, the incentives may have been insufficiently salient to move behavior, as they were introduced only once, at the start of the study. This second limitation bears directly on our pre-registered hypothesis. Because that hypothesis concerned whether stakes would raise suspension and whether AI would blunt this effect, a weak or easily-forgotten incentive could, in principle, explain both the small stakes effect and the absent interaction. Study 3 was designed to rule this out.

Study 3 ($N = 843$) therefore conceptually replicated Study 2 while addressing these limitations. To make the incentive salient and continuously active, participants were reminded of the stakes before every question; to remove the possibility that eliciting confidence encouraged guessing, confidence judgments were dropped. See *Methods* for details.

Consistent with the previous study, access to AI advice substantially reduced judgment suspension both in the absence of monetary stakes (0.32 vs. 0.02; $b = -0.300$, $SE = 0.027$, $t = -11.03$, $p < .001$) and when monetary stakes were present (0.41 vs. 0.08; $b = -0.329$, $SE = 0.032$, $t = -10.37$, $p < .001$). See *Tables S9-S10* for statistical details. Thus, even with more salient incentives and no confidence elicitation, the availability of AI advice nearly eliminated judgment suspension.

Unlike Study 2, monetary stakes here significantly increased judgment suspension, both when AI advice was unavailable (0.32 vs. 0.41; $b = 0.089$, $SE = 0.039$, $t = 2.296$, $p = .022$) and when it was available (0.02 vs. 0.08; $b = 0.062$, $SE = 0.016$, $t = 3.911$, $p < .001$). The pre-registered AI × stakes interaction was again not significant ($b = -0.028$, $SE = 0.042$, $t = -0.665$, $p = .506$; see *Figure 3a*). See *Tables S9-S10* for statistical details. Thus, stakes raised judgment suspension by a similar amount whether or not AI was present, rather than having their effect blunted by AI as we had predicted.

Monetary incentives also altered how participants responded to AI advice. Correctness analyses revealed a significant interaction between stakes and AI advice ($b = 0.063$, $SE = 0.028$, $t = 2.233$, $p = .026$). Like Study 2, when AI advice was available, monetary stakes increased correctness (0.11 vs. 0.16; $b = 0.055$, $SE = 0.017$, $t = 3.164$, $p = .002$). By contrast, when AI advice was unavailable, stakes had no effect on correctness (0.28 vs. 0.27; $b = -0.009$, $SE = 0.022$, $t = -0.382$, $p = .703$; see *Figure 3b*). See *Tables S11-S12* for statistical details. This pattern indicates that incentives improved performance specifically when participants had the opportunity to use AI advice.

Importantly, Study 3 replicated the behavioral pattern observed in Study 2. Participants sought AI advice substantially less frequently when stakes were present than when they were absent (4.53 vs. 5.27 times out of six questions; $b = -0.744$, $SE = 0.182$, $t = -4.083$, $p < .001$; see *Figure 3c*).

Thus, even with salient, per-question incentives, stakes raised judgment suspension without their effect being blunted by AI—ruling out weak incentives as the explanation for the null interaction in Study 2, and indicating that AI availability and stakes contribute independently.

Together, Studies 2–3 suggest that, while monetary incentives do not restore judgment suspension to the levels observed when AI advice is unavailable, they nonetheless make participants less willing to rely on AI advice and more likely to provide correct answers.

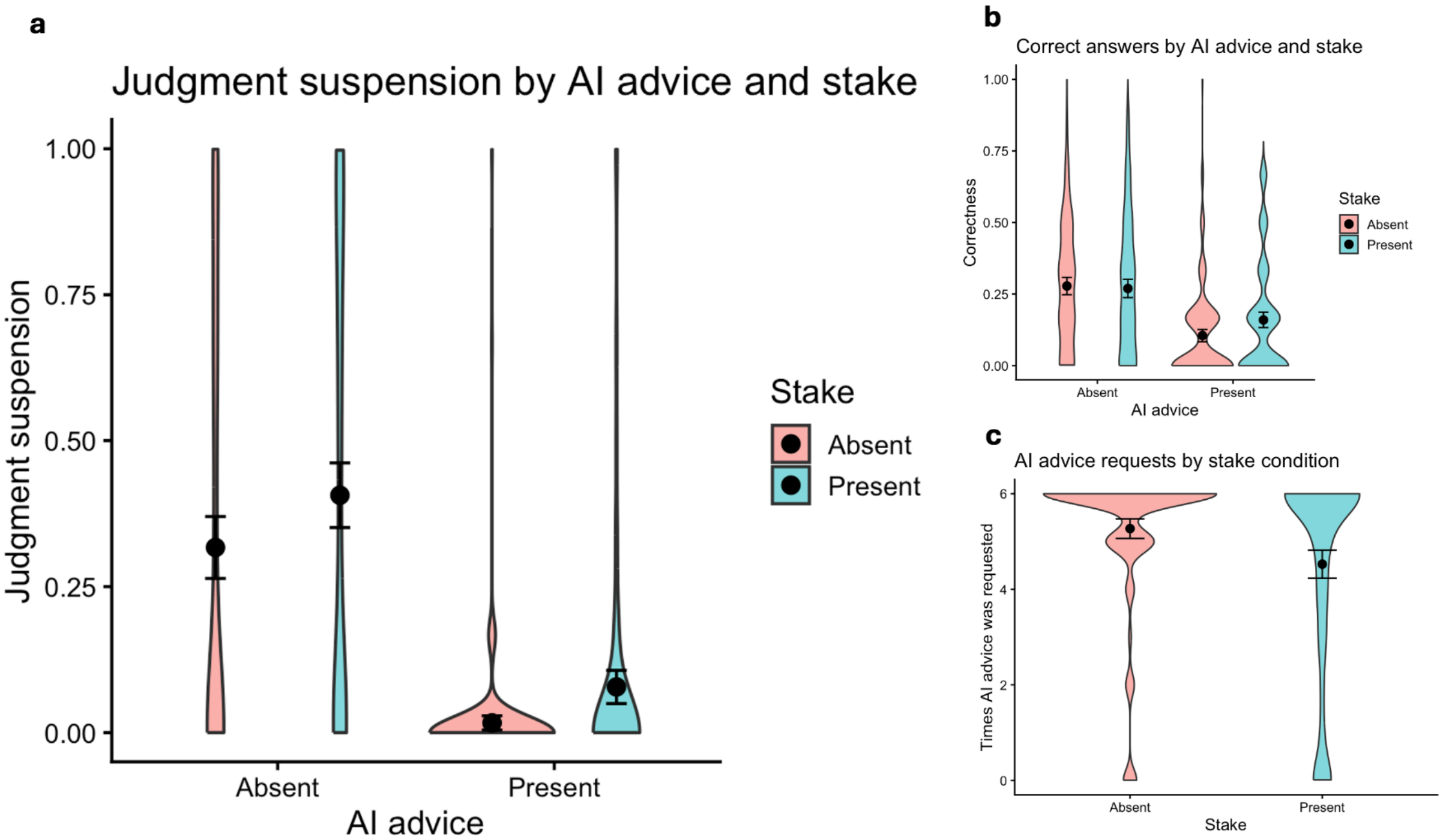

**Fig. 3 | Effects of AI advice and monetary stakes on judgment suspension, correctness, and AI-advice requests in Study 3.** The interaction between stakes and AI advice was not significant: monetary stakes increased judgment suspension both when AI advice was absent and when it was present (**a**). Correctness analyses revealed a significant interaction between stakes and AI advice: monetary stakes increased the likelihood of correct answers in the AI-advice condition relative to the no-stake condition (**b**). Participants requested AI advice substantially less frequently when monetary stakes were present (**c**). Points represent condition means and error bars indicate 95% confidence intervals.

The previous studies raise a theoretically and practically important question: what happens when people cannot avoid AI advice? Outside the laboratory, AI recommendations are often presented automatically rather than actively requested. Search engines increasingly provide AI-generated summaries by default, writing assistants offer unsolicited suggestions, and educational and workplace software may display AI recommendations before users have formed their own judgments.

Study 4 ($N = 853$) therefore removed participants' control over whether to receive AI advice. It was identical to Study 3 except that, in the conditions where AI advice was available, it appeared automatically rather than on request. See *Methods* for details.

The results largely replicated those of Study 3. Access to AI advice continued to dramatically reduce judgment suspension, both when monetary stakes were absent (0.35 vs 0.01; $b = -0.333$, $SE = 0.029$, $t = -11.56$, $p < 0.001$) and when they were present (0.39 vs 0.07; $b = -0.322$, $SE = 0.032$, $t = -10.18$, $p < 0.001$). Monetary stakes increased judgment suspension when AI advice was present (0.01 vs. 0.07; $b = 0.060$, $SE = 0.016$, $t = 3.692$, $p < .001$) and tended to increase judgment suspension when AI advice was absent, although not significantly so (0.35 vs. 0.39; $b = 0.048$, $SE = 0.041$, $t = 1.182$, $p = .238$). The interaction between stakes and AI advice was again not significant ($b = 0.012$, $SE = 0.043$, $t = 0.274$, $p = .784$; see *Figure 4a*). See *Tables S13-S14* for statistical details.

Across Studies 2–4, then, our pre-registered prediction of a negative AI × stakes interaction on suspension was not supported ($p = .274$, .506, and .784, respectively). AI availability and stakes acted largely independently: AI sharply reduced suspension regardless of stakes, while stakes modestly increased suspension—most clearly in Study 3—regardless of AI. That incentives left the AI effect on suspension essentially intact underscores its strength; rather than moderating the deferral to AI, consequences acted alongside it. The one place stakes and AI interacted was accuracy, to which we turn next.

Monetary incentives again altered participants' behavior when AI advice was available. Correctness analyses revealed a significant interaction between stakes and AI advice ($b = 0.072$, $SE = 0.028$, $t = 2.565$, $p = .011$). When AI advice was available, monetary stakes increased correctness (0.07 vs. 0.14; $b = 0.072$, $SE = 0.016$, $t = 4.420$, $p < .001$). By contrast, when AI advice was unavailable, monetary stakes had no meaningful effect on correctness (0.27 vs. 0.27; $b = -0.000$, $SE = 0.023$, $t = -0.012$, $p = .991$; see *Figure 4b*). See *Tables S15-S16* for statistical details. Thus, again, monetary incentives increase people's accuracy specifically when AI advice is unavailable.

Taken together, Studies 2–4 suggest that although monetary incentives do not restore judgment suspension to the levels observed without AI advice, they improve accuracy when AI is available by making participants more likely to override that advice.

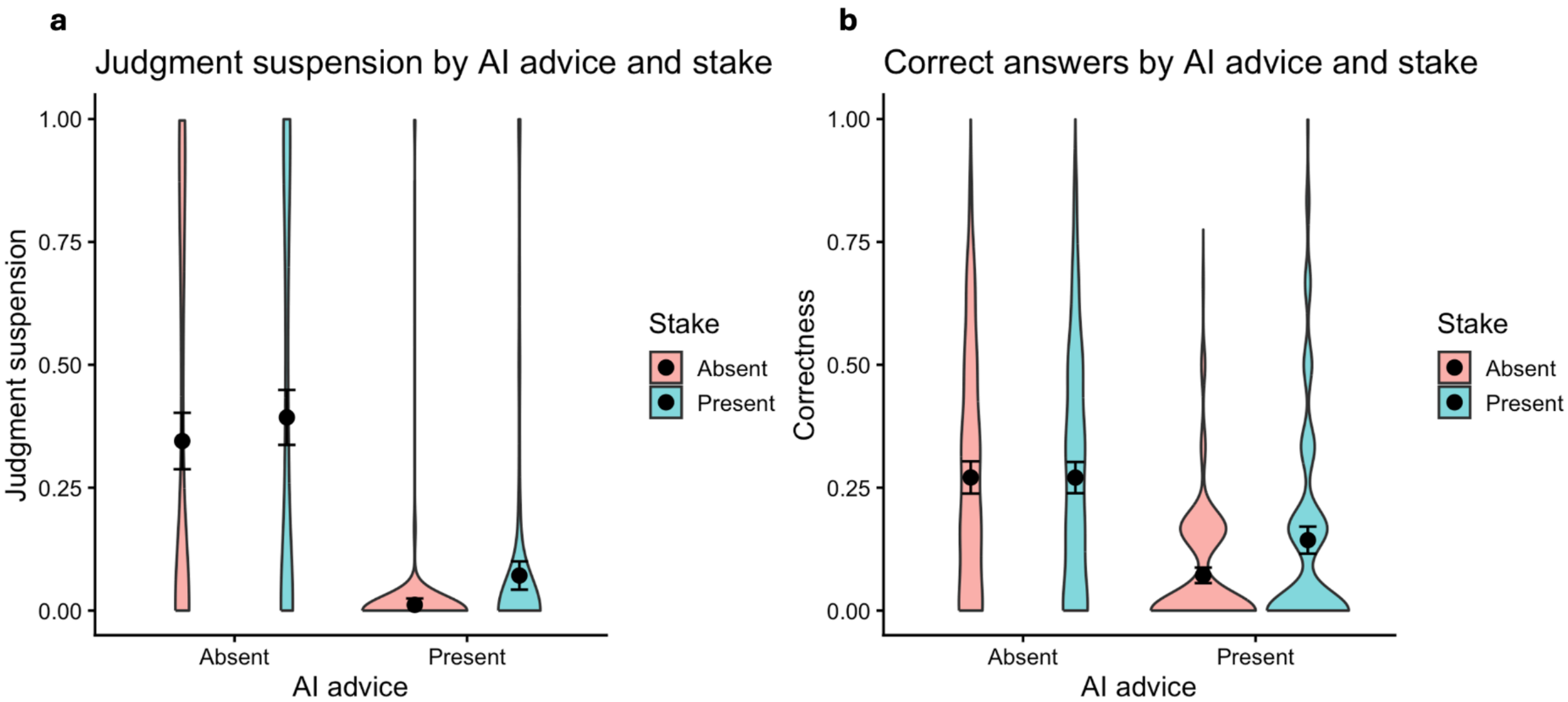

**Fig. 4 | Effects of AI advice and monetary stakes on judgment suspension and correctness in Study 4.** The interaction between stakes and AI advice was not significant (**a**). Correctness analyses revealed a significant interaction between stakes and AI advice. Monetary stakes increased correctness only when AI advice was provided (**b**) . Points represent condition means and error bars indicate 95% confidence intervals.

## Discussion

Across five experiments, mere access to AI advice nearly eliminated participants' willingness to suspend judgment, even though that advice was usually wrong. Because we engineered the questions so that the AI typically erred, this collapse cannot be read as sensible delegation to a reliable tool. Participants abandoned "I don't know" in favor of answers that were more often incorrect, while reporting higher confidence. Specifically, in the absence of monetary incentives, participants answered more questions but were correct about a third as often as when AI was unavailable (pooled correctness: 27.5% vs. 9.2%), while confidence was roughly two and a half times as high (mean confidence: 29.6 vs. 75.9). Without incentives, AI access made people far more assured and far less accurate.

Monetary stakes did not moderate the AI effect on suspension—contrary to our pre-registered prediction, the AI × stakes interaction was non-significant in all three studies—but stakes exerted independent effects. When accuracy was rewarded, participants sought AI advice less, suspended judgment somewhat more, and answered more accurately; yet suspension remained far below the level observed when AI was unavailable, whether advice was optional (Studies 2–3) or displayed by default (Study 4). Two conclusions follow. First, the near-elimination of suspension under AI access is not a fixed cognitive limitation but a contextual response that monetary consequences partly correct: people can recruit their own judgment when motivated, so what AI changes is willingness, not capacity. Second, monetary stakes left a large residual effect, indicating that the pull toward deferring to AI is strong enough to survive small incentives.

These findings can be read through the notion of *epistemia* (Loru et al., 2025; Quattrociocchi et al., 2025): the tendency to accept AI outputs for their surface plausibility and syntactic coherence rather than through verification. Humans can monitor uncertainty and withhold a response when warranted, whereas an LLM must always produce an output and does not itself suspend judgment under ignorance (Kalai et al., 2026; Strack, 2026); when users delegate judgment to such a system, they may inherit its lack of suspension, and linguistic fluency may substitute for epistemic evaluation. One interpretation, consistent with our data, is that the availability of a fluent answer lowers the threshold at which people decide they know enough to respond—reducing suspension, inflating confidence, and, because the advice was usually wrong, lowering accuracy. Stakes are one condition under which this mechanism partly re-engages.

This degree of deference to AI advice is striking against what is known about advice use. People typically *underweight* others' advice, shifting only about a third of the way toward an advisor when averaging would serve accuracy better—an egocentric discounting that ordinarily protects independent judgment (Bonaccio & Dalal, 2006). Our participants did the opposite, and two features make this especially notable. First, access to AI did not merely fill gaps in knowledge: in the no-stakes conditions the AI group answered more questions than the baseline group yet produced fewer correct ones, so AI advice must have converted some would-be-correct responses into errors—people abandoned answers they knew, or could have reached, for a confident but wrong suggestion. Second, an advice that diverges from one's own tentative belief is a conflicting cue, and conflicting cues normally lower confidence (Koriat, 2012) and should, if anything, increase withholding (Koriat & Goldsmith, 1996); instead, the divergent and usually incorrect AI advice reduced suspension and raised confidence, as though the fluent response resolved deliberation prematurely rather than being weighed against the participant's own view.

The moderating role of stakes links these results to automation bias and complacency—over-reliance on automated aids, with less scrutiny as trust accrues (Parasuraman & Manzey, 2010). Because holding people accountable for accuracy reduces such over-reliance and the commission errors that follow from accepting a recommendation despite disconfirming evidence (Skitka, Mosier, & Burdick, 2000), our stakes manipulation has a natural parallel: rewarding accuracy led participants to rely on the AI less and override it more often. We did not record whether they verified the advice elsewhere, so we cannot speak to the vigilance mechanism emphasised in that work. Yet our data isolate the resulting choice. Suspension offers a complementary index of over-reliance, reflecting whether users engaged their own uncertainty.

The confidence findings deserve emphasis: confidence nearly doubled even as accuracy fell. This mirrors evidence that access to external information inflates self-assessed knowledge. Searching the Internet leads people to overrate how well they can explain things (Fisher, Goddu, & Keil, 2015). Similarly outside assistance breeds overconfidence more generally (Fisher & Oppenheimer, 2021), usually understood through transactive memory, in which people conflate a partner's knowledge with their own (Sparrow, Liu, & Wegner, 2011). Our results extend this from information that genuinely exists online to generative advice that is factually wrong: participants drew confidence from a partner whose answers were wrong, and did so even though suspension remained available. The risk is arguably sharper than with search, because a generative model returns a single, decisive answer where a search might surface conflicting sources, removing the friction that could prompt a pause.

More broadly, our findings speak to when an LLM acts as a benign external cognitive artefact and when it supplants cognition (Clark & Chalmers, 1998; Koskinen, 2024). Casati (2017) distinguishes a mode in which a function is delegated to and taken over by the artefact from one in which the artefact supports a function without overtaking it. Our results suggest that stakes help decide which obtains—whether AI advice replaces judgment or merely informs it. The critical question may thus be not whether AI is used but under what conditions users retain independent metacognitive control.

Several limitations qualify these conclusions. First, all studies used a single class of stimuli—fine visual details in films—chosen precisely because they reliably induce hallucinations. We see no reason the effect should be specific to this content: the mechanism we propose operates on the availability and fluency of an answer, not on its subject matter, so any domain in which an AI readily supplies a confident response should elicit the same lowering of the threshold to answer. Consistent with this, the effect held across questions ranging from widely seen films to obscure titles, suggesting it does not depend on familiarity with the specific material. What genuinely remains open is the boundary condition of AI reliability: our questions were engineered so the AI was almost always wrong, and it is untested whether suspension collapses to the same degree when AI advice is usually correct. Testing domains where errors carry real-world rather than monetary consequences is a further priority. Second, our design did not record whether participants who overrode the AI consulted other sources, such as the open Internet or another model, which limits inferences about the strategies behind successful overrides. Future work should directly measure verification behavior to distinguish between reduced reliance on AI and increased use of alternative information sources. Third, our incentives, though effective, were modest; whether larger or reputational stakes would close more of the gap to baseline is unknown.

These results open several directions. One concerns AI literacy: whether teaching users how and when LLMs fail can restore judgment suspension and, more generally, guard against cognitive surrender (Shaw & Nave 2026; Meincke et al. 2026; Quattrociocchi et al., 2025; Voinea et al., 2026). Another concerns mechanism: whether suspension collapses because users trust the AI, because an available answer lowers the perceived cost of committing, or because fluency short-circuits the metacognitive signal that would otherwise trigger a pause. Distinguishing between these mechanisms will be essential for understanding whether the observed effect reflects changes in trust, effort allocation, metacognitive monitoring, or a combination of these processes.

As AI-generated answers become ubiquitous and, increasingly, unsolicited, our results show that the willingness to say "I don't know" may be among the first casualties of human–AI interaction. Whether human judgment keeps its footing as AI systems spread may ultimately depend less on making AI more accurate than on preserving people's readiness to recognize, and to act on, the limits of what they know.

## Methods

### *Questions common across studies*

In every study, participants were asked to answer, where possible, the same six questions about films. The questions were designed to be both hard for participants to answer and likely to induce hallucinations (i.e., fabricated answers) from the LLM used in our experiments, Step 3.5 Flash:

1) What animal is on the bow of the pirate ship from "Asterix and Obelix"?
2) In the movie "The Grand Budapest Hotel", what is Agatha's signature hairstyle?
3) What color is the team's uniform in "Bend It like Beckham"?
4) What vehicle does Monica drive in "Like a Cat on a Highway"?
5) What color is the turtle in the animated movie "Momo" by Enzo d'Alò?
6) What pet animal does Asenath have in "Joseph King of Dreams"?

In the Supplementary Information, Section S1.1, we report several examples of hallucinations from Step 3.5 Flash as well as from several state-of-art LLMs, including ChatGPT-5.5 and Claude Sonnet 4.6.

***Study 1a***

**Participants.** An a priori power analysis (one-tailed $t$-test, $\alpha = 0.05$, power = 0.99) indicated that detecting a medium effect ($d = 0.5$) would require 254 participants. We set recruitment to 300 on Prolific to allow for dropouts and exclusions. The final dataset contained 317 complete responses: a number of participants completed the survey in Qualtrics without registering their submission on Prolific, so the Qualtrics export exceeded the Prolific target. Three IP addresses appeared more than once; for each, we retained the first submission (by start time) and excluded the duplicate. We analyzed the remaining 314 complete responses.

**Design.** Participants were randomly assigned to one of two conditions. In the AI condition, participants could consult an LLM (StepFun's Step 3.5 Flash) while answering; in the baseline condition, no LLM was available. All participants answered the same six questions (see *Questions common across studies*) and could decline to answer at any point. We selected this LLM because it was straightforward to embed in Qualtrics. Its name was never shown to participants, and we have no reason to expect that the specific LLM used influenced participants' behaviour.

**Procedure.** For participants in the AI condition, each question was accompanied by the option to submit it to the AI tool. Consulting the AI was optional. Selecting this option prompted Qualtrics to send the question to Step 3.5 Flash with an instruction to answer briefly, and to display the model's response. Participants then entered their own answer. Participants in the baseline condition answered without the option of seeking AI advice. The base Prolific compensation was £0.30. Full experimental instructions are available on the OSF page of this project: https://osf.io/nwx6r/overview?view_only=a03ae0ca587c4c84a442c44d55ab97b6.

**Measure.** Our pre-registered outcome was *judgment suspension*, defined as the proportion of responses indicating a withholding of judgment (e.g., "I don't know", "I'm not sure", or a blank response), out of the six questions per participant.

**Coding.** Judgment-suspension responses were identified by manual coding by two of the authors. The coders first agreed on a classification rule: blank responses, explicit expressions of ignorance (e.g., "I don't know"), and non-responsive entries such as random typing were coded as judgment suspension, whereas any response containing a substantive attempt at an answer, including tentative or hedged guesses (e.g., "green?"), was not. Applying this rule, agreement between the two coders was nearly complete; the few discrepancies reflected isolated coding errors by one coder and were resolved through discussion. We had planned to validate this coding with two large language models (GPT-5.5 and Claude Opus 4.8), but both proved unreliable on this classification task, so we did not pursue it.

**Hypothesis.** We pre-registered the directional hypothesis that participants in the AI condition would show lower judgment suspension than those in the baseline condition.

**Analysis.** We compared judgment suspension between conditions using a one-tailed independent-samples t-test, consistent with the pre-registered directional hypothesis.

**Pre-registration.** The pre-registration for this study is available at: https://aspredicted.org/g3bn74.pdf

***Study 1b***

**Participants.** As in Study 1a, we aimed at recruiting 300 participants. Participants were recruited via Prolific, restricted to US-based respondents, and screened to exclude anyone who had taken part in Study 1a. As before, the final dataset slightly exceeded the target (again owing to responses completed in Qualtrics but not registered on Prolific), yielding 315 complete responses. Five ProlificIDs appeared more than once; for each, we retained the first submission (by start time) and excluded the duplicate. We analyzed the remaining 310 complete responses.

**Design.** The design matched Study 1a: participants were randomly assigned to an AI condition or a baseline condition, answered the same six questions (see *Questions common across studies*), and could decline to answer at any point. The two studies differed only in how AI advice was delivered.

**Procedure.** In the AI condition, each question was again accompanied by the option to consult the AI. However, rather than querying the model live, selecting this option displayed one of three pre-recorded responses, drawn at random, that Step 3.5 Flash had generated for that question during Study 1a (see *SI*, Section S1.2). We adopted this fixed-response procedure to ensure that every participant who sought advice received a complete, well-formed answer, eliminating the live-query failures and latency that can occur when the model is under heavy load. Because all displayed answers were genuine Step 3.5 Flash outputs, the advice participants saw was equivalent in kind to that in Study 1a, but its delivery was held constant across participants. The baseline condition was identical to Study 1a. Full experimental instructions are reported in the OSF page of this article. The base Prolific compensation was £0.30.

**Measure.** As in Study 1a, the main outcome was *judgment suspension*, defined as the proportion of responses indicating a withholding of judgment.

**Coding.** As in the previous study, judgment-suspension responses were identified through manual coding by two of the authors, following the same classification rule; agreement was again near-total, and the few discrepancies were resolved by discussion.

**Hypothesis.** We again hypothesized that participants in the AI condition would show lower judgment suspension than those in the baseline condition.

**Analysis.** We compared judgment suspension between conditions using a one-tailed independent-samples t-test, consistent with our hypothesis.

**Pre-registration.** This study was not pre-registered, because it is almost identical to Study 1a.

***Study 2***

**Participants.** We conducted an a priori power analysis in G*Power for an $F$-test of a single predictor within a linear multiple regression (fixed model, $R^2$ increase; 1 tested predictor, 3 predictors total). To detect a small interaction effect ($f^2 = 0.02$) at $\alpha = 0.05$ with 95% power, the analysis indicated a required sample of $N = 648$. We set recruitment to 800 on Prolific (200 per condition) to allow for dropouts and exclusions, restricted participation to US-based respondents, and screened out anyone who had taken part in Studies 1a or 1b. As in the earlier studies, the final dataset exceeded the target, yielding 831 complete responses. Nineteen ProlificIDs appeared more than once; for each, we retained the first submission (by start time) and excluded the duplicate. We analyzed the remaining 812 complete responses.

**Design.** Participants were randomly assigned to one of four conditions in a 2 (AI advice: available vs. unavailable) × 2 (stakes: present vs. absent) between-subjects design. In the two AI conditions, participants could consult Step 3.5 Flash while answering; in the two no-AI conditions, no LLM was available. In the two stakes conditions, accuracy was incentivised: participants earned \$0.10 for each correct answer, lost \$0.10 for each incorrect answer, and received \$0 for suspending judgment. In the two no-stakes conditions, responses carried no monetary consequence. All participants answered the same six questions (see *Questions common across studies*) and could decline to answer at any point.

**Procedure.** As in Study 1b, participants in the AI conditions saw, for each question, the option to consult the AI; selecting it displayed one of three pre-registered responses, drawn at random, that Step 3.5 Flash had generated for that question in Study 1a. After every question, all participants reported how confident they were in their answer on a continuous 0–100 scale ("How confident are you about your answer?"). In the stakes conditions, the payment rule was stated explicitly before the task. The base Prolific compensation was £0.30. Exact experimental instructions are reported in the OSF page of this article.

**Measures.** The primary dependent variable was *judgment suspension*, the individual-level frequency of responses indicating a withholding of judgment. Secondary variables were *correctness* (the frequency of correct answers) and *confidence* (the mean rating on the 0–100 scale).

**Coding.** Judgment suspension and correctness were scored by manual coding by two authors. Correct answers are provided in the SI, Section S1.1. As in the previous studies, agreement was near-total, and the few discrepancies were resolved by discussion.

**Hypotheses.** Our primary, pre-registered hypothesis was that the effect of stakes on judgment suspension would be attenuated when AI was available, corresponding to a negative coefficient on the AI × stakes interaction. As a secondary hypothesis, we predicted that in the absence of AI, stakes would increase judgment suspension, corresponding to a positive coefficient on the stakes term.

**Analysis.** We pre-registered a linear regression predicting judgment suspension from an AI indicator (1 = AI available, 0 = not), a stakes indicator (1 = stakes present, 0 = absent), and their interaction. Exploratory analyses with correctness and confidence as dependent variables are conducted using structurally the same statistical test.

**Pre-registration.** The pre-registration for this study is available at: https://aspredicted.org/wu5fs2.pdf

***Study 3***

**Participants.** The power analysis was identical to Study 2: an a priori G*Power analysis for an $F$-test of a single predictor in a linear multiple regression (fixed model, $R^2$ increase; 1 tested predictor, 3 predictors total) indicated that detecting a small interaction effect ($f^2 = 0.02$) at $\alpha = 0.05$ with 95% power required $N = 648$. We set recruitment to 800 on Prolific (200 per condition) to allow for dropouts and exclusions, restricted participation to US-based respondents, and screened out anyone who had taken part in Studies 1a, 1b, or 2. As in the earlier studies, the final dataset exceeded the target, yielding 862 complete responses. Nineteen ProlificIDs appeared more than once; for each, we retained the first submission (by start time) and excluded the duplicate. We analyzed the remaining 843 complete responses.

**Design.** The design replicated Study 2: a 2 (AI advice: available vs. unavailable) × 2 (stakes: present vs. absent) between-subjects design, with the same incentive structure in the stakes conditions ($0.10 per correct answer, −$0.10 per incorrect answer, $0 for suspending judgment). As in Studies 1b and 2, AI responses were not generated live but drawn at random from a fixed set of pre-recorded Step 3.5 Flash outputs. All participants answered the same six questions (see *Questions common across studies*) and could decline to answer at any point. Study 3 differed from Study 2 in two respects: it did not collect confidence ratings, and it made incentives more salient by reminding participants of the stakes on every question.

**Procedure.** As in Study 2, participants in the AI conditions saw, for each question, the option to consult the AI; selecting it displayed one of three pre-registered responses, drawn at random, that Step 3.5 Flash had generated for that question in Study 1a. In the two stakes conditions, each question was accompanied by an on-screen reminder of the payment rule. The base Prolific compensation was £0.30. Exact experimental instructions are reported in the OSF page of this article.

**Measures.** The primary dependent variable was *judgment suspension*, the individual-level frequency of responses indicating a withholding of judgment. The secondary variable was *correctness*, the individual-level frequency of correct answers.

**Coding.** Judgment suspension and correctness were scored by manual coding by two authors. As in the previous studies, agreement was near-total, and the few discrepancies were resolved by discussion.

**Hypotheses.** As in Study 2, our primary pre-registered hypothesis was that the effect of stakes on judgment suspension would be attenuated when AI was available, corresponding to a negative coefficient on the AI × stakes interaction. As a secondary hypothesis, we predicted that in the absence of AI, stakes would increase judgment suspension, corresponding to a positive coefficient on the stakes term.

**Analysis.** We pre-registered a linear regression predicting judgment suspension from an AI indicator (1 = AI available, 0 = not), a stakes indicator (1 = stakes present, 0 = absent), and their interaction. The exploratory analysis with correctness as dependent variable is conducted using structurally the same statistical test.

**Pre-registration.** The pre-registration for this study is available at: https://aspredicted.org/re9sq6.pdf

***Study 4***

**Participants.** The power analysis was identical to Studies 2 and 3: an a priori G*Power analysis for an $F$-test of a single predictor in a linear multiple regression (fixed model, $R^2$ increase; 1 tested predictor, 3 predictors total) indicated that detecting a small interaction effect ($f^2 = 0.02$) at $\alpha = 0.05$ with 95% power required $N = 648$. We set recruitment to 800 on Prolific (200 per condition) to allow for dropouts and exclusions, restricted participation to US-based respondents, and screened out anyone who had taken part in Studies 1a, 1b, 2, or 3. As in the earlier studies, the final dataset exceeded the target, yielding 879 complete responses. Twenty-six ProlificIDs appeared more than once; for each, we retained the first submission (by start time) and excluded the duplicate. We analyzed the remaining 853 complete responses.

**Design.** The design replicated Study 3: a 2 (AI advice: present vs. absent) × 2 (stakes: present vs. absent) between-subjects design with the same incentive structure in the stakes conditions ($0.10 per correct answer, −$0.10 per incorrect answer, $0 for suspending judgment). There was one key change. In the previous studies, participants in the AI conditions *chose* whether to consult the model; in Study 4, AI advice was instead displayed automatically alongside every question, removing participants' control over whether to receive it. As in Studies 1b, 2, and 3, the advice shown was not generated live but drawn at random from a fixed set of pre-recorded Step 3.5 Flash outputs. All participants answered the same six questions (see *Questions common across studies*) and could decline to answer at any point.

**Procedure.** For participants in the two AI conditions, each question was presented together with one of three pre-registered responses, drawn at random, that Step 3.5 Flash had generated for that question in Study 1a; the response appeared automatically, without any action by the participant. In the two stakes conditions, each question was accompanied by an on-screen reminder of the payment rule. The base Prolific compensation was £0.30. Full experimental instructions are reported in the Supplementary Information.

**Measures.** The primary dependent variable was *judgment suspension*, the individual-level frequency of responses indicating a withholding of judgment. The secondary variable was *correctness*, the individual-level frequency of correct answers.

**Coding.** Judgment suspension and correctness were scored by manual coding by two authors. As in the previous studies, agreement was near-total, and the few discrepancies were resolved by discussion.

**Hypotheses.** As in Studies 2 and 3, our primary pre-registered hypothesis was that the effect of stakes on judgment suspension would be attenuated when AI was available, corresponding to a negative coefficient on the AI × stakes interaction. As a secondary hypothesis, we predicted that in the absence of AI, stakes would increase judgment suspension, corresponding to a positive coefficient on the stakes term.

**Analysis.** We pre-registered a linear regression predicting judgment suspension from an AI indicator (1 = AI present, 0 = absent), a stakes indicator (1 = stakes present, 0 = absent), and their interaction. The exploratory analysis with correctness as dependent variable is conducted using structurally the same statistical test.

**Pre-registration.** The pre-registration for this study is available at: https://aspredicted.org/bf92yw.pdf

**Data and code availability.** All data, code, preregistrations, and experimental instructions are available at: https://osf.io/nwx6r/overview?view_only=a03ae0ca587c4c84a442c44d55ab97b6

**Competing interests.** The authors declare no competing interests.
**Author contributions.** CM, WQ, VC designed the studies. CM, VC analysed the data. CM and VC wrote the original draft. CM, WQ and VC reviewed and edited the manuscript.

## Supplementary Information

### Section S1. Information about how LLMs respond to the questions we selected for this work

### Section S1.1. Several leading state-of-art LLMs hallucinate

Here is an example of answers provided by three of the main commercial language models, ChatGPT-5.5, Gemini and Claude Sonnet 4.6, as of July 9th 2026:

*Question 1*: What animal is on the bow of the pirate ship from “Asterix and Obelix”?

> *ChatGPT-5.5*: 🏴‍☠️ The Pirate Ship Figurehead in *Asterix & Obelix*
> *[Two images were displayed here, which we did not include for reasons of copyrights. The first, a frame from the Asterix and Obelix comics, displaying the pirate captain on the bird-shaped bow of his ship. The second, a picture of a toy ship from the Asterix and Obelix merchandise, with a bird-shaped bow.]*
> The animal on the bow (figurehead) of the pirate ship in *Asterix & Obelix* is typically depicted as a stylized sea monster or dragon-like creature.
> 🐉 Why a dragon/sea monster?
> - The pirates in the series (inspired by classic caricatures of pirates like Redbeard) sail a ship with exaggerated, humorous details.
> - The figurehead resembles a mythical creature, often with an open mouth and fierce expression—more decorative than realistic.
> - This fits the comic’s playful tone, blending historical elements with fantasy and parody.
>
> 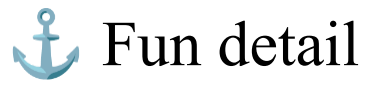
> ⚓ Fun detail
> The pirate ship and its crew are recurring gag characters in the series—they almost always end up shipwrecked after crossing paths with Asterix and Obelix 😄
> If you want, I can show how the design changes across different albums or films!
>
> *Gemini 3.5 Flash*: The figurehead on the bow of the unfortunate pirates' ship in *Asterix and Obelix* is a gargoyle-like dragon.
> Because their ship is heavily inspired by a Viking longship (drakkar), it features a stylized, green-colored dragon head on the prow and a matching tail at the stern. Of course, since they have the terrible luck of constantly running into the Gauls, that dragon figurehead usually ends up floating in the water alongside the wreckage!
>
> *Claude Sonnet 4.6*:

*[Three images of toys from the Asterix and Obelix merchandise were displayed here, which we did not include for reasons of copyrights. Two of them show the pirates on a raft, one shows them on a ship with a bow shaped like a bird.]*
There isn't one single, consistent animal on the bow of the pirates' ship in *Asterix* — that's actually part of the running joke. According to Asterix's official site, Albert Uderzo drew a whole "hilarious collection of pirate-ship bows" across the various albums, changing the figurehead design from book to book rather than sticking to one fixed emblem.
So depending on which album or adaptation you're looking at, the figurehead varies — sometimes a mythical creature, sometimes something more whimsical — rather than there being one canonical animal tied to Captain Redbeard's ship. If you have a specific album or film in mind, I can look up what's drawn on the bow in that particular one.

*Correct answer*: A sea monster or a griffin. *Acceptable answers:* A monster. / A sea monster. / A griffin. / A dragon. / A bird.

*Question 2*: In the movie "The Grand Budapest Hotel", what is Agatha's signature hairstyle?

*ChatGPT-5.5*: 🎀 Agatha's Signature Hairstyle in *The Grand Budapest Hotel*
*[Three images of Agatha from the movie were displayed here, which we did not include for reasons of copyrights.]*
Agatha, played by Saoirse Ronan in *The Grand Budapest Hotel*, is known for her milkmaid braids—a hairstyle where two braids are wrapped around the top of the head like a crown.
💡 Key features of her look:

- Center-parted hair, neatly divided into two sections
- Braids pinned over the crown, forming a halo-like shape
- A soft, tidy style that gives her a wholesome, vintage, storybook charm

🎬 Why it stands out:
The hairstyle perfectly matches the film's whimsical, old-European aesthetic and reinforces Agatha's gentle, romantic character. It's simple, iconic, and instantly recognizable.

*Gemini 3.5 Flash*: Agatha's signature hairstyle in *The Grand Budapest Hotel* is a crown braid (also called a milkmaid braid). To perfectly match Wes Anderson's meticulous aesthetic, her hair is parted down the middle and tightly braided into a crown around her head, which beautifully frames her face—and highlights her birthmark in the shape of Mexico.

*Claude Sonnet 4.6*: Agatha's signature hairstyle in *The Grand Budapest Hotel* is a pair of milkmaid braids — twin braids crossed and pinned around her head like a crown. One writer described her as "all milkmaid braids and the silhouette of Mexico upon her face," referencing her matching birthmark. Fan costume guides consistently point to the same feature: Agatha's ginger milkmaid braids are considered one of the most iconic hairstyles among Wes Anderson's characters, right alongside Suzy's updo in *Moonrise Kingdom* and Margot Tenenbaum's hair clips.
For DIY versions, fans often add a small wheat stalk tucked into the braid and a light dusting of flour, nodding to her job as a baker's apprentice at Mendl's.

*Correct answers*: A crown braid. / A milkmaid braid. / A german braid. / A Heidi braid. / A halo braid. *Acceptable answer:* A braid going around her head.

*Question 3*: What color is the team's uniform in "Bend It like Beckham"?

*ChatGPT-5.5*: ⚽ Team Uniform Color in *Bend It Like Beckham*
*[Three images from the movie were displayed here, which we did not include for reasons of copyrights. They show the two protagonists wearing white uniforms with red details.]*
In *Bend It Like Beckham*, the women's team (the Hounslow Harriers) wears red uniforms, typically paired with white accents.
🔴 Details:

- Primary color: Red
- Secondary color: White (for stripes, trim, or shorts)
- The bold red kit helps make the team stand out on the field and adds to the film's energetic, sporty vibe.

If you're thinking of a specific scene or match, I can break down any variations too!

*Gemini 3.5 Flash*: The uniform for the Hounslow Harriers, the women's football team in *Bend It Like Beckham*, is red and white.
The players wear bright red jerseys (sponsored by Adidas) paired with white shorts and red socks.

*Claude Sonnet 4.6*: The Hounslow Harriers — the (fictional) team Jess and Jules play for — wear a white and red kit: white Adidas shirts with red trim/shorts. Multiple sources confirm this, including original production jerseys sold as memorabilia, an Adidas football shirt in white and red produced for use in the film, paired with red shorts.
One fan interviewed about the film's style specifically called out the classic white and red kits the team wears as an iconic look, noting how rare it is to see such stripped-back, classic kits these days.

*Correct answers*: White and red. *Acceptable answers:* White. / Red.

*Question 4*: What vehicle does Monica drive in “Like a Cat on a Highway”?

*ChatGPT-5.5*: Monica, the character played by Paola Cortellesi in Like a Cat on a Highway (Italian: *Come un gatto in tangenziale*), is most often associated with a small scooter/moped, which she uses to get around Rome's outskirts and later for pizza deliveries after opening her pizzeria. The film emphasizes the contrast between Monica's modest, working-class lifestyle and Giovanni's more affluent urban life.
If you're asking about a specific scene, there are also moments involving cars (including Giovanni's car, whose windshield Monica famously smashes with a baseball bat), but Monica's characteristic personal vehicle is a scooter.

*Gemini 3.5 Flash*: In the Italian comedy *Like a Cat on a Highway* (*Come un gatto in tangenziale*), Monica (played by Paola Cortellesi) does not actually drive a car of her own.
Instead, she notoriously relies on public transportation (taking the bus) or relies on others for rides. A running plot point in the movie revolves around travel and transit—such as the chaotic bus trip her son Alessio and Giovanni's daughter Agnese take to get to the crowded beach of *Coccia di Morto*, starkly contrasting with Giovanni's comfortable life and his Volvo SUV.
However, when Monica and her family do travel by car, it is usually in the Fiat Multipla driven by her ex-husband, Sergio (Claudio Amendola).

*Claude Sonnet 4.6*: According to the Internet Movie Car Database, the vehicle used extensively by a main character in the film is a 2016 Toyota Auris Touring Sports Hybrid — that's Monica's car. It's rated with four stars, indicating it's a vehicle used a great deal by a main character throughout the film.

*Step 3.5 Flash*: In the movie *Like a Cat on a Highway*, Monica drives a small, older-model Fiat 500. It is a 2007 Fiat 500, and the car is a notable part of her character's practical and resourceful nature during the chaotic road trip.

*Correct answer*: A 2008 blue Toyota Aygo. *Acceptable answers:* A car. / A Toyota. / An old car. / A blue car. / An old Toyota. / A blue Toyota./ A 2008 Toyota. / A Toyota Aygo.

*Question 5*: What color is the turtle in the animated movie “Momo” by Enzo d’Alò?

*ChatGPT-5.5*: 🐢 The Turtle in *Momo* (2001)
*[Three images from the movie and what looks like other internet resources were displayed here, which we did not include for reasons of copyrights. They show a green turtle with a brown shell.]*

In *Momo alla conquista del tempo*, the turtle named Cassiopée is generally depicted in green tones.
🎨 Details:

- Her shell and body are shown in green (often slightly muted or earthy green)
- The design sometimes includes subtle shading toward olive or green-brown, fitting the film's soft, storybook color palette

Although descriptions of the film mainly mention her role rather than her appearance (Première), the animation itself consistently portrays her as a green turtle, which aligns with typical visual conventions for turtles in animation.
👉 So, the short answer: green.

*Gemini 3.5 Flash*: *[Three images from the movie were displayed here, which we did not include for reasons of copyrights. They show the turtle with its shell looking alternatively brown and blue.]*
In Enzo d'Alò's animated film *Momo alla conquista del tempo* (*Momo*), Cassiopea the magical turtle has a distinct light green body with a dark green and yellow/orange patterned shell.
When she communicates by making words light up directly on her shell, the glowing text shines in a bright magic blue/green or golden-yellow hue depending on the scene.

*Claude Sonnet 4.6*: Based on the images, Cassiopea (the tortoise in Enzo d'Alò's *Momo*) has a purple/violet-toned shell with a lighter, pale skin tone — she's often depicted with a distinctive lavender-purple carapace rather than a typical green or brown turtle color, which fits the film's stylized, dreamlike visual palette (inspired by artists like Dalí and Magritte).

*Correct answer*: Green body with a brown shell which glows blue when using magical powers. *Acceptable answers:* Green / Brown / Blue

*Question 6*: What pet animal does Asenath have in "Joseph King of Dreams"?

*ChatGPT-5.5*: 🐱 Asenath's Pet in *Joseph: King of Dreams*
In *Joseph: King of Dreams*, Asenath has a pet cat, named Akia.
🐾 Details:

- The animal is clearly shown in the film's early scenes, where Asenath and others are chasing and trying to calm the cat. (Dreamworks Animation)
- The cat plays a small but memorable role—Joseph even wins Asenath over by calming it with a bowl of milk. (Fun Trivia)
- The pet's name, Akia, is explicitly identified in character information sources. (universalstudios.fandom.com)

👉 Answer: A cat (named *Akia*).

*Gemini 3.5 Flash*: In the DreamWorks film *Joseph: King of Dreams*, Asenath has a pet **cat** named **Akia** (sometimes spelled Akeila).
She is first introduced in the movie while chasing the cat with her sisters, which leads to her first interaction with Joseph when he calms the animal down with a bowl of milk.

*Claude Sonnet 4.6*: Asenath has a pet cat (named Akia).
She is first introduced in the film when she and her sisters are chasing the cat, and Joseph impresses her by successfully calming it down with a bowl of milk.

*Correct answer*: A grey-brown, possibly Egyptian, cat named Akia. *Acceptable answers:* A cat / A grey cat / A brown cat / An Egyptian cat / A cat named Akia.

**Section S1.2. Answers from Step 3.5 Flash used in Studies 1b, 2-4**

The following sets of responses from Step 3.5 Flash were recorded from Study 1a and used as pre-registered responses in the following studies. Correct answers at the end of each set.

*Question 1*: What animal is on the bow of the pirate ship from "Asterix and Obelix"?

*Suggestion 1*: The animal on the bow of the pirate ship from the Asterix and Obelix movies is a boar. This is a recurring feature in the series, where the pirate ship captained by Bloody Beard has a boar's head as its figurehead

*Suggestion 2*: The animal on the bow of the pirate ship in the movie "Asterix & Obelix: Mission Cleopatra" is a crocodile. This figurehead appears during the pirate attack scene and ties into the film's Egyptian setting.

*Suggestion 3*: The animal on the bow of the pirate ship in the Asterix and Obelix stories, including the movies, is typically a wolf. This figurehead appears on the ship of the recurring pirate captain, who is often humorously defeated by the Gauls.

*Correct answer*: A sea monster or a griffin. *Acceptable answers:* A monster / A sea monster / A griffin / A dragon / A bird.

*Question 2*: In the movie "The Grand Budapest Hotel", what is Agatha's signature hairstyle?

*Suggestion 1*: In the movie “Grand Budapest Hotel”, Agatha's signature hairstyle is finger waves. Her dark hair is styled in neat, close-to-the-head S-shaped waves, often parted in the center, which was a popular look in the 1930s setting of the film.

*Suggestion 2*: In the movie “The Grand Budapest Hotel”, Agatha's signature hairstyle consists of two braids that are wrapped into circular buns or loops. These are positioned on either side of her head, just above her ears, and are a distinctive part of her appearance as a pastry chef at the hotel.

*Suggestion 3*: In the movie “The Grand Budapest Hotel”, Agatha's signature hairstyle is soft, loose waves that fall just past her shoulders, often styled with a gentle side part and sometimes tucked behind one ear. The most distinctive feature associated with her appearance is the pentagram-shaped birthmark on her neck, but her hairstyle itself is a simple, romantic wave.

*Correct answers*: A crown braid / A milkmaid braid / A german braid / A Heidi braid / A halo braid. *Acceptable answer:* A braid going around her head.

*Question 3*: What color is the team’s uniform in “Bend It like Beckham”?

*Suggestion 1*: The Hustlers football team, which Jess joins in the movie “Bend It Like Beckham”, wears blue uniforms.

*Suggestion 2*: In the movie “Bend It Like Beckham”, the team Jess joins, the Hounslow Harriers, wears a bright yellow uniform. Their home kit consists of yellow shirts with black trim and black shorts.

*Suggestion 3*: The main football team in the movie “Bend it like Beckham”, the Hounslow Harriers, wears white uniforms with dark blue trim. Their shirts are white with blue details, and they wear dark shorts.

*Correct answers*: White and red. *Acceptable answers:* White / Red.

*Question 4*: What vehicle does Monica drive in “Like a Cat on a Highway”?

*Suggestion 1*: In the movie “Like a Cat on a Highway”, Monica drives a small, older-model Fiat 500. It is a 2007 Fiat 500, and the car is a notable part of her character's practical and resourceful nature during the chaotic road trip.

*Suggestion 2*: In the movie “Like a Cat on a Highway”, Monica, played by Monica Bellucci, drives an Alfa Romeo Giulia Quadrifoglio.

*Suggestion 3*: Monica drives a small, colorful vintage Fiat 600. Specifically, it is a 1960 Fiat 600 Multipla in the movie.

*Correct answer*: A 2008 blue Toyota Aygo. *Acceptable answers:* A car / A Toyota / An old car / A blue car / An old Toyota / A blue Toyota / A 2008 Toyota / A Toyota Aygo.

*Question 5*: What color is the turtle in the animated movie “Momo” by Enzo d’Alò?

*Suggestion 1*: The turtle in the animated movie “Momo” by Enzo d’Alò is grey. It is often referred to as the Grey Turtle, based on the original novel by Michael Ende, upon which the film is adapted.

*Suggestion 2*: The turtle in the animated movie “Momo” by Enzo d'Alò is yellow with black patterns on its shell.

*Suggestion 3*: The turtle in the animated movie “Momo” by Enzo d'Alò is green. Its shell and skin are both shades of green, consistent with the natural coloration of many turtle species.

*Correct answer*: Green body with a brown shell which glows blue when using magical powers. *Acceptable answers:* Green / Brown / Blue.

*Question 6*: What pet animal does Asenath have in “Joseph King of Dreams”?

*Suggestion 1*: In the animated movie “Joseph: King of Dreams”, Asenath has a small pet monkey. It is a female capuchin monkey that she keeps as a companion.

*Suggestion 2*: In the animated film “Joseph: King of Dreams”, Asenath has a pet falcon. She is frequently shown caring for it and it appears with her in several scenes.

*Suggestion 3*: Asenath has a pet dove in the movie “Joseph King of Dreams”. The dove appears with her throughout the story, symbolizing peace and hope.

*Correct answer*: A grey-brown, possibly Egyptian, cat named Akia. *Acceptable answers:* A cat / A grey cat / A brown cat / An Egyptian cat / A cat named Akia.

## Section S2. Detailed statistical results

### Section S2.1. Studies 1a-1b

**Table S1.** Judgment suspension by condition in Studies 1a and 1b.

| Study | M (No AI) | M (AI) | Mean diff. | *t* | *df* | 95% CI | *p* |
|---|---|---|---|---|---|---|---|
| **1a** | 0.357 | 0.064 | 0.293 | 8.295 | 197.92 | [0.235, ∞) | $8.25 \times 10^{-15}$ |
| **1b** | 0.436 | 0.026 | 0.410 | 11.772 | 170.36 | [0.342, ∞) | $< 2.2 \times 10^{-16}$ |

*Note.* The dependent variable is the per-participant proportion of responses indicating judgment suspension (0–1). "No AI" denotes the baseline condition and "AI" the condition with access to AI advice. Statistics are from Welch's two-sample *t*-tests, one-tailed, testing the pre-registered directional hypothesis that judgment suspension is lower under AI access; the reported 95% confidence intervals are correspondingly one-sided, [lower bound, ∞).

### Section S2.2. Study 2.

**Table S2.** Judgment suspension by AI advice and stakes (Study 2): cell means, simple effects, and their interaction.

| | No AI | AI | Simple effect of AI |
|---|---|---|---|
| **No stakes** | 0.173 | 0.014 | b = −0.159<br>t(390) = −6.81, p < .001 |
| **Stakes** | 0.211 | 0.018 | b = −0.193<br>t(418) = −7.96, p < .001 |
| **Simple effect of stakes** | b = 0.038<br>t(394) = 1.43, p = .253 | b = 0.004<br>t(414) = 0.39, p = .700 | **AI × Stakes**<br>b = −0.034<br>t(808) = −1.01, p = .312 |

*Note.* Central entries are mean judgment suspension (proportion of responses, 0–1) in each cell of the 2 × 2 design (N = 831). Marginal entries are simple effects estimated by linear regression within each level of the other factor: the effect of AI advice at each stakes level (right column) and the effect of stakes at each AI level (bottom row), each reported as the unstandardized coefficient *b* with its *t* statistic (df in parentheses) and *p* value. Simple effects use within-subset error terms. The lower-right entry is the AI × Stakes interaction from the full factorial model.

**Table S3.** Linear regression predicting judgment suspension from AI advice, stakes, and their interaction (Study 2).

| Predictor | *b* | SE | *t* | *p* |
|---|---|---|---|---|
| Intercept | 0.172 | 0.017 | 9.96 | $< 2 \times 10^{-16}$ |
| AI advice | −0.159 | 0.024 | −6.55 | $1.04 \times 10^{-10}$ |
| Stakes | 0.038 | 0.024 | 1.57 | .117 |
| AI advice × Stakes | −0.034 | 0.034 | −1.01 | .312 |

*Note.* The dependent variable is the per-participant proportion of responses indicating judgment suspension (0–1). AI advice (1 = available, 0 = not) and Stakes (1 = present, 0 = absent) are dummy-coded; the intercept corresponds to the no-AI, no-stakes condition. Model fit: $R^2 = .122$, adjusted $R^2 = .119$, residual SE = 0.240, $F(3, 808) = 37.38$, $p < .001$.

**Table S4.** Response confidence by AI advice and stakes (Study 2): cell means, simple effects, and their interaction.

| | No AI | AI | Effect of AI |
|---|---|---|---|
| **No stakes** | 29.6 | 75.9 | b = 46.38<br>t(386) = 17.01, p < .001 |
| **Stakes** | 37.7 | 73.5 | b = 35.87<br>t(409) = 13.46, p < .001 |
| **Effect of stakes** | b = 8.08<br>t(381) = 2.56, p = .011 | b = −2.43<br>t(414) = −1.10, p = .264 | **AI × Stakes**<br>b = −10.51<br>t(795) = −2.76, p = .006 |

*Note.* Central entries are mean response confidence (0–100 scale) in each cell of the 2 × 2 design (N = 799; 13 observations excluded for missingness). The "Effect of AI" column reports simple effects of AI availability estimated at each stakes level (unstandardized *b*, *t* with df in parentheses, *p*). The "Effect of stakes" row reports simple effects of stakes estimated within each AI level. Simple effects use within-subset error terms. The lower-right entry is the AI × Stakes interaction from the full model.

**Table S5.** Linear regression predicting response confidence from AI advice, stakes, and their interaction (Study 2).

| **Predictor** | ***b*** | **SE** | ***t*** | ***p*** |
|---|---|---|---|---|
| Intercept | 29.57 | 1.97 | 15.02 | $< 2 \times 10^{-16}$ |
| AI advice | 46.38 | 2.74 | 16.96 | $< 2 \times 10^{-16}$ |
| Stakes | 8.08 | 2.75 | 2.94 | .003 |
| AI advice × Stakes | −10.51 | 3.81 | −2.76 | .006 |

*Note.* N = 799 (13 observations excluded for missingness). The dependent variable is response confidence on a 0–100 scale. AI advice (1 = available, 0 = not) and Stakes (1 = present, 0 = absent) are dummy-coded; the intercept corresponds to the no-AI, no-stakes condition. Model fit: $R^2 = .372$, adjusted $R^2 = .370$, residual SE = 26.972 $F(3, 795) = 157.3$, $p < .001$.

**Table S6.** Proportion of correct answers by AI advice and stakes (Study 2): cell means, simple effects, and their interaction.

| | No AI | AI | Effect of AI |
|---|---|---|---|
| **No stakes** | 0.276 | 0.100 | b = −0.176<br>t(390) = -11.40, p < .001 |
| **Stakes** | 0.334 | 0.173 | b = −0.161<br>t(418) = -7.437, p < .001 |

| **Effect of stakes** | b = 0.058<br>t(394) = 2.84, p = .005 | b = 0.074<br>t(414) = 4.17, p < .001 | **AI × Stakes**<br>b = 0.015<br>t(808) = 0.56, p = .576 |
|---|---|---|---|

*Note.* Central entries are the proportions of correct answers, 0–1, in each cell of the 2 × 2 design (N = 812). The "Effect of AI" column reports the AI–no-AI difference in accuracy at each stakes level (both $p < .001$). The "Effect of stakes" row reports simple effects of stakes estimated within each AI level (unstandardized $b$, $t$ with df in parentheses, $p$). Simple effects use within-subset error terms. The lower-right entry is the AI × Stakes interaction from the full model.

**Table S7.** Linear regression predicting accuracy (proportion correct) from AI advice, stakes, and their interaction (Study 2).

| Predictor | *b* | SE | *t* | *p* |
|---|---|---|---|---|
| Intercept | 0.276 | 0.014 | 19.86 | $< 2 \times 10^{-16}$ |
| AI advice | −0.176 | 0.019 | −9.09 | $< 2 \times 10^{-16}$ |
| Stakes | 0.058 | 0.019 | 3.02 | .003 |
| AI advice × Stakes | 0.015 | 0.027 | 0.56 | .576 |

*Note.* The dependent variable is the per-participant proportion of correct answers (0–1). AI advice (1 = available, 0 = not) and Stakes (1 = present, 0 = absent) are dummy-coded; the intercept corresponds to the no-AI, no-stakes condition. Model fit: $R^2 = .182$, adjusted $R^2 = .180$, residual SE = 0.192, $F(3, 808) = 60.3$, $p < .001$.

**Table S8.** Linear regression predicting AI-advice seeking from stakes, within the AI-available group.

| Predictor | *b* | SE | *t* | *p* |
|---|---|---|---|---|
| Intercept (no stakes) | 5.438 | 0.115 | 47.33 | $< 2 \times 10^{-16}$ |
| Stakes | −0.503 | 0.160 | −3.15 | .002 |

*Note.* Estimated within the AI-available conditions only (N = 415). The dependent variable is the number of questions (0–6) on which a participant requested AI advice. Stakes is dummy-coded (1 = present, 0 = absent); the intercept is the no-stakes mean. Model fit: $R^2 = .023$, adjusted $R^2 = .021$, residual SE = 1.629, $F(1, 414) = 9.90$, $p = .002$.

## Section S2.3. Study 3

**Table S9.** Linear regression predicting judgment suspension from AI advice, stakes, and their interaction (Study 3).

| Predictor | *b* | SE | *t* | *p* |
|---|---|---|---|---|
| Intercept | 0.317 | 0.021 | 14.95 | $< 2 \times 10^{-16}$ |
| AI advice | −0.300 | 0.030 | −10.09 | $< 2 \times 10^{-16}$ |
| Stakes | 0.089 | 0.030 | 3.01 | .003 |
| AI advice × Stakes | −0.028 | 0.042 | −0.67 | .506 |

*Note.* N = 843. The dependent variable is the per-participant proportion of responses indicating judgment suspension (0–1). AI advice (1 = available, 0 = not) and Stakes (1 = present, 0 = absent) are dummy-coded; the intercept corresponds to the no-AI, no-stakes condition. Model fit: $R^2 = .222$, adjusted $R^2 = .220$, residual SE = 0.304, $F(3, 839) = 80.13$, $p < .001$.

**Table S10.** Judgment suspension by AI advice and stakes (Study 3): cell means, simple effects, and their interaction.

| | **No AI** | **AI** | **Effect of AI** |
|---|---|---|---|
| **No stakes** | 0.317 | 0.017 | b = −0.300<br>t(414) = −11.03, p < .001 |
| **Stakes** | 0.407 | 0.078 | b = −0.329<br>t(425) = −10.37, p < .001 |
| **Effect of stakes** | b = 0.089<br>t(417) = 2.30, p = .022 | b = 0.062<br>t(422) = 3.91, p < .001 | **AI × Stakes**<br>b = −0.028<br>t(839) = −0.67, p = .506 |

*Note.* Central entries are mean judgment suspension (proportion of responses, 0–1) in each cell of the 2 × 2 design (N = 843). Marginal entries are simple effects estimated by linear regression within each level of the other factor: the effect of AI advice at each stakes level (right column) and the effect of stakes at each AI level (bottom row), each as the unstandardized coefficient *b* with its *t* statistic (df in parentheses) and *p* value. The lower-right entry is the AI × Stakes interaction from the full factorial model.

**Table S11.** Linear regression predicting accuracy (proportion correct) from AI advice, stakes, and their interaction (Study 3).

| **Predictor** | ***b*** | **SE** | ***t*** | ***p*** |
|---|---|---|---|---|
| Intercept | 0.278 | 0.014 | 19.42 | $< 2 \times 10^{-16}$ |
| AI advice | −0.173 | 0.020 | −8.60 | $< 2 \times 10^{-16}$ |
| Stakes | −0.009 | 0.020 | −0.43 | .669 |
| AI advice × Stakes | 0.063 | 0.028 | 2.23 | .026 |

*Note.* N = 842 (1 observation excluded for missingness). The dependent variable is the per-participant proportion of correct answers (0–1). AI advice (1 = available, 0 = not) and Stakes (1 = present, 0 = absent) are dummy-coded; the intercept corresponds to the no-AI, no-stakes condition. Model fit: $R^2 = .114$, adjusted $R^2 = .111$, residual SE = 0.205, $F(3, 838) = 35.88$, $p < .001$.

**Table S12.** Accuracy (proportion correct) by AI advice and stakes (Study 3): cell means, simple effects, and their interaction.

| | **No AI** | **AI** | **Effect of AI** |
|---|---|---|---|
| **No stakes** | 0.278 | 0.105 | b = −0.173<br>t(414) = −9.24, p < .001 |
| **Stakes** | 0.269 | 0.160 | b = −0.110<br>t(424) = −5.20, p < .001 |

| Effect of stakes | b = −0.009<br>t(417) = −0.38, p = .703 | b = 0.055<br>t(421) = 3.16, p < .001 | **AI × Stakes**<br>b = 0.066<br>t(859) = 2.38, p = .018 |
|---|---|---|---|

*Note.* Central entries are mean accuracy (proportion of correct answers, 0–1) in each cell of the 2 × 2 design (N = 863). Marginal entries are simple effects estimated by linear regression within each level of the other factor (unstandardized *b*, *t* with df in parentheses, *p*).

## Section S2.4. Study 4

**Table S13.** Linear regression predicting judgment suspension from AI advice, stakes, and their interaction (Study 4).

| Predictor | *b* | SE | *t* | *p* |
|---|---|---|---|---|
| Intercept | 0.345 | 0.022 | 15.59 | $< 2 \times 10^{-16}$ |
| AI advice | −0.333 | 0.031 | −10.86 | $< 2 \times 10^{-16}$ |
| Stakes | 0.048 | 0.031 | 1.56 | .119 |
| AI advice × Stakes | 0.012 | 0.043 | 0.27 | .784 |

*Note.* N = 853. In Study 4, AI advice was displayed automatically rather than requested. The dependent variable is the per-participant proportion of responses indicating judgment suspension (0–1). AI advice (1 = present, 0 = absent) and Stakes (1 = present, 0 = absent) are dummy-coded; the intercept corresponds to the no-AI, no-stakes condition. Model fit: $R^2$ = .221, adjusted $R^2$ = .218, residual SE = 0.313, F(3, 849) = 80.14, *p* < .001.

**Table S14.** Judgment suspension by AI advice and stakes (Study 4): cell means, simple effects, and their interaction.

| | **No AI** | **AI** | **Effect of AI** |
|---|---|---|---|
| **No stakes** | 0.345 | 0.012 | b = −0.333<br>t(414) = −11.56, p < .001 |
| **Stakes** | 0.393 | 0.071 | b = −0.322<br>t(435) = −10.18, p < .001 |
| **Effect of stakes** | b = 0.048<br>t(413) = 1.18, p = .238 | b = 0.060<br>t(436) = 3.69, p < .001 | **AI × Stakes**<br>b = 0.012<br>t(849) = 0.27, p = .784 |

*Note.* Central entries are mean judgment suspension (proportion of responses, 0–1) in each cell of the 2 × 2 design (N = 853). Marginal entries are simple effects estimated by linear regression within each level of the other factor (unstandardized *b*, *t* with df in parentheses, *p*).

**Table S15.** Linear regression predicting accuracy (proportion correct) from AI advice, stakes, and their interaction (Study 4).

| Predictor | *b* | SE | *t* | *p* |
|---|---|---|---|---|
| Intercept | 0.271 | 0.014 | 18.72 | $< 2 \times 10^{-16}$ |
| AI advice | −0.199 | 0.020 | −9.92 | $< 2 \times 10^{-16}$ |
| Stakes | −0.000 | 0.020 | −0.01 | .989 |
| AI advice × Stakes | 0.072 | 0.028 | 2.57 | .011 |

*Note.* N = 853. The dependent variable is the per-participant proportion of correct answers (0–1). AI advice (1 = present, 0 = absent) and Stakes (1 = present, 0 = absent) are dummy-coded; the intercept corresponds to the no-AI, no-stakes condition. Model fit: $R^2 = .149$, adjusted $R^2 = .146$, residual SE = 0.205, $F(3, 849) = 49.36$, $p < .001$.

**Table S16.** Accuracy (proportion correct) by AI advice and stakes (Study 4): cell means, simple effects, and their interaction.

| | **No AI** | **AI** | **Effect of AI** |
|---|---|---|---|
| **No stakes** | 0.271 | 0.072 | b = −0.199<br>t(414) = −11.03, p < .001 |
| **Stakes** | 0.271 | 0.143 | b = −0.127<br>t(435) = −5.97, p < .001 |
| **Effect of stakes** | b = −0.000<br>t(413) = −0.01, p = .991 | b = 0.072<br>t(436) = 4.42, p < .001 | **AI × Stakes**<br>b = 0.072<br>t(849) = 2.57, p = .011 |

*Note.* Bold entries are mean accuracy (proportion of correct answers, 0–1) in each cell of the 2 × 2 design (N = 876); in Study 4, AI advice was displayed automatically rather than requested. Marginal entries are simple effects estimated by linear regression within each level of the other factor (unstandardized *b*, *t* with df in parentheses, *p*).